\documentclass{article}

\usepackage{arxiv}

\usepackage[utf8]{inputenc} 
\usepackage[T1]{fontenc}    
\usepackage{hyperref}       

\usepackage{nicefrac}       
\usepackage{microtype}      
\usepackage{lipsum}		

\usepackage[square, numbers]{natbib}
\usepackage{doi}

\usepackage{bm}
\usepackage{algorithm}
\usepackage{amsmath,amsfonts}
\usepackage{amssymb}
\usepackage{array}
\usepackage{booktabs}
\usepackage{changepage}
\usepackage{color}
\usepackage{graphicx}
\usepackage{makecell}
\usepackage{url}
\usepackage{siunitx}
\usepackage{multirow}
\usepackage{mathtools}
\usepackage{pifont}       
\newcommand\Plus{\texttt{+}}

\title{You Don't Need Attention: Gated Convolutional Modeling for Watch-Based Fall Detection}

\author{
  Sana Alamgeer$^1$ \\
  \texttt{sanaalamgeer@gmail.com} \\
  \And
  Ronish Kumar$^1$ \\
  \texttt{omc46@txstate.edu} \\
  \And
  Awatif Yasmin$^1$ \\
  \texttt{nuc4@txstate.edu} \\
  \And
  Muhammad Irshad$^1$ \\
  \texttt{muhammadirshad@txstate.edu} \\ 
  \And
  Anne H. H. Ngu$^1$ \\
  \texttt{angu@txstate.edu} \\ 
  \AND
  $^1$Texas State University, San Marcos, USA
}

\date{}

\renewcommand{\shorttitle}{Gated-CNN for Watch-Based Fall Detection}

\begin{document}
\maketitle

\begin{abstract}
Existing deep learning approaches for wearable fall detection systems rely on self-attention mechanisms that impose quadratic computational overhead, distributing weights across all time steps. This global weight distribution impairs the precise localization of the brief impact signatures that characterize falls within short, fixed-length windows. 
To overcome this challenge, we propose Gated-CNN, a lightweight dual-stream architecture that processes accelerometer and gyroscope streams through independent one-dimensional convolutional feature extractors, followed by (i) a sigmoid gating module that selectively suppresses uninformative background activations while amplifying fall-discriminative features, (ii) a global average pooling layer that compresses each stream into a compact fixed-length descriptor, and (iii) a shared classification head that fuses both descriptors for binary fall prediction. 
For offline evaluation, we evaluate the model across five wrist-mounted inertial measurement unit (IMU) datasets, achieving average F1-scores of 93\%, 93\%, 90\%, 91\%, and 90\% on SmartFallMM, WEDA-Fall, FallAllD, UMAFall, and UP-Fall, outperforming Transformer baselines. For real-time evaluation, we deployed the model on a Google Pixel Watch 3 and tested across 12 participants. The model achieves an average F1-score of 97\% and an accuracy of 98\% with zero missed falls, showing that sigmoid gating offers a more structurally aligned and computationally efficient alternative to attention for commodity smartwatch-based fall detection.
\end{abstract}


\section{Introduction}\label{sect:introduction}

\begin{figure}
 \centering
 \includegraphics[width=0.8\textwidth]{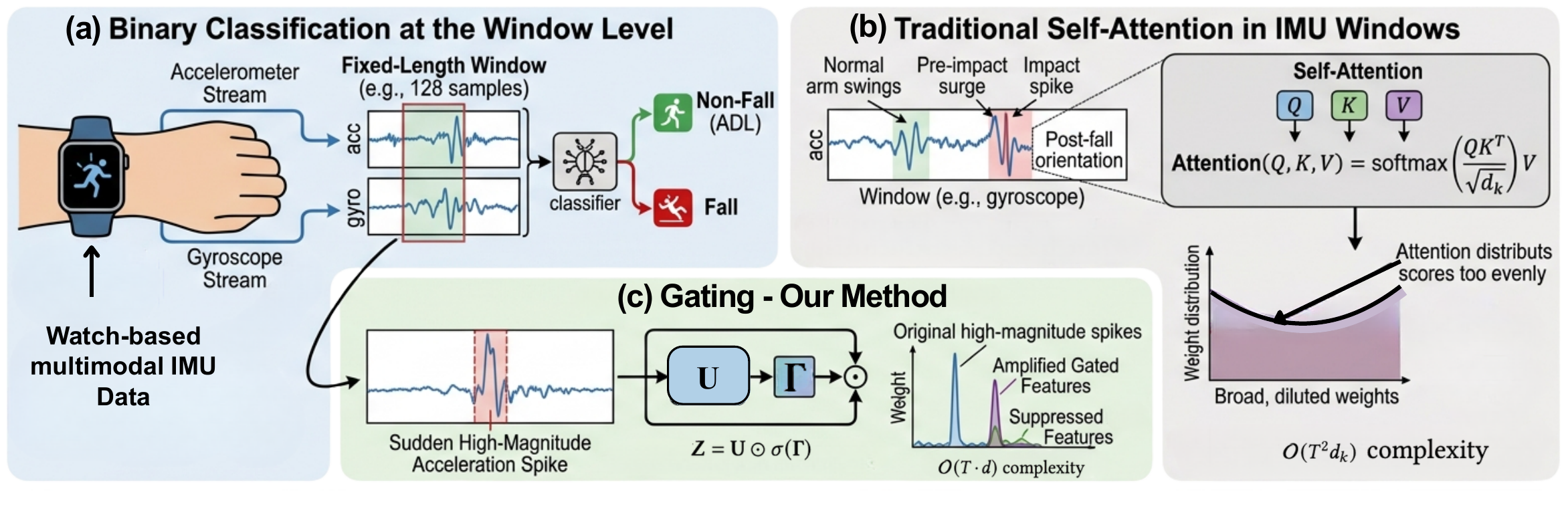}
 \caption{Overview of fall detection problem: (a) Window-level binary classification, (b) Self-attention distributes weights broadly across all time steps with $\mathcal{O}(T^2 d_k)$ complexity. (c) Gating (our method) selectively amplifies the high-magnitude impact spike and suppresses surrounding uninformative motion at $\mathcal{O}(T \cdot d)$ complexity.}
 \label{fig:intro}
\end{figure}

Smartwatches equipped with inertial measurement units (IMUs) are now widely adopted for health and activity monitoring in daily life. This has made watch-based fall detection a practical and unobtrusive approach to continuous safety monitoring~\cite{Gu2026WearableSensors, Gattani2026}. In such systems, as illustrated in Fig.~\ref{fig:intro}(a), fall detection is formulated as a binary classification problem, wherein a sliding window of fixed length ($64$--$128$ samples) is extracted from the continuous multimodal IMU signals, comprising both accelerometer and gyroscope measurements from a single wrist-mounted device, and classified as either a fall or a non-fall event~\cite{Marques2023, Xinyao2025}. This window-level paradigm is well-suited for real-time deployment on resource-constrained devices. However, it introduces a fundamental constraint: the model must produce a reliable binary decision from a compact, fixed-length temporal segment that may capture only a partial or ambiguous snapshot of the fall dynamics. 

To enhance the model's ability to reason over such short temporal segments, the self-attention~\cite{vaswani2017attention} has been widely integrated into fall detection and human activity recognition (HAR) pipelines~\cite{Zafar2025, 11089122, s25175249}. As shown in Fig.~\ref{fig:intro}(b), in these methods, for each time step $t$, a query vector $\mathbf{Q}_t$ (what this step is looking for), a key vector $\mathbf{K}_t$ (what this step is about), and a value vector $\mathbf{V}_t$ (what this step actually contains), are computed from the input feature sequence, and  and the output is a weighted sum of all value vectors:
\begin{equation}
    \text{Attention}(\mathbf{Q}, \mathbf{K}, \mathbf{V}) = \text{softmax} \left( \frac{\mathbf{Q}\mathbf{K}^{\top}}{\sqrt{d_k}}\right)\mathbf{V},
\end{equation}
where $\mathbf{Q}, \mathbf{K}, \mathbf{V} \in \mathbb{R}^{T \times d_k}$, $T$ is the sequence length, $d_k$ is the key dimensionality, and ${(\cdot)^\top}$ denotes the matrix transpose. While this mechanism is effective for long-range dependency modeling in natural language processing (NLP) and video understanding, its application to window-level HAR and fall detection reveals several concrete limitations. First, the pairwise dot-product between all positions incurs a computational complexity of $\mathcal{O}(T^2 d_k)$, which introduces unnecessary overhead for short fixed-length IMU windows. Second, the softmax normalization forces the attention weights to sum to one across all $T$ positions. As a result, truly discriminative segments, such as the impact phase of a fall, have their contributions diluted by the non-zero weights assigned to uninformative or noisy time steps~\cite{Mudarisov2025}. 

Gating mechanisms address these limitations more directly and with lower computational cost. Inspired by Gated Linear Units (GLU)~\cite{glu2017}, a gating module decomposes the input feature tensor $\mathbf{F} \in \mathbb{R}^{T \times d}$ into two parallel projections, a contextual projection $\mathbf{U}$ (meaningful patterns) and a gate $\mathbf{\Gamma}$ (deciding how much of each motion feature in $\mathbf{U}$ is relevant and should be kept), and computes their element-wise product:
\begin{equation}
    \mathbf{Z} = \mathbf{U} \odot \sigma(\mathbf{\Gamma}), \quad  \mathbf{U} = \phi(\mathbf{W}_u \mathbf{F}), \quad  \mathbf{\Gamma} = \mathbf{W}_g \mathbf{F}
\end{equation}
where $\mathbf{W}_u, \mathbf{W}_g \in \mathbb{R}^{d \times d}$ are learnable projection matrices, $\phi(\cdot)$ is a non-linear activation function, $\sigma(\cdot)$ is the sigmoid activation function, and $\odot$ denotes element-wise multiplication. Unlike attention, gating operates at 
$\mathcal{O}(T \cdot d)$ complexity, linear in both sequence length and feature dimensionality, and imposes no constraint that the weights sum to one across positions. Each feature activation is independently scaled by a value in $(0, 1)$, allowing the module to fully suppress uninformative activations to near zero while preserving or amplifying discriminative ones. In the context of window-level fall detection, as shown in Fig.~\ref{fig:intro}(c), this behavior is particularly advantageous: the abrupt kinematic signature of a fall, typically confined to a small fraction of the window, can be selectively retained while surrounding non-informative steady-state motion is attenuated, without the dilution effect introduced by softmax normalization in attention. 

Building on these advantages of gating, this work proposes a lightweight dual-stream architecture that processes accelerometer and gyroscope signals from a single wrist-mounted device through parallel, independent branches, each consisting of successive one-dimensional convolutional neural network (1D-CNN) layers followed by a GLU-inspired gating module and a global average pooling (GAP) layer. The two stream descriptors are concatenated and passed to a shared classification head for binary fall prediction. To the best of our knowledge, this is the first work to employ gating as a refinement module within a fall detection pipeline, offering a lightweight and structurally aligned alternative to attention for window-level binary classification.

The remainder of this paper is organized as follows. Section~\ref{sect:related_work} reviews related work on wrist-based fall detection and feature refinement techniques. Section~\ref{sect:methodology} presents the proposed architecture in detail. Sections~\ref{sect:impl_details} and~\ref{sect:results_discussion} detail the experimental setup and results, respectively. Finally, Section~\ref{sect:conclusion} presents the conclusions and future directions.

\section{Related Work}\label{sect:related_work}
In this section, we review three categories of deep learning-based approaches for wrist-based fall detection: recurrent networks, self-attention-based transformers, and gating mechanisms, with contributions and limitations of each.

\subsection{Recurrent Networks for Fall Detection}

Early deep learning approaches to wrist-based fall detection relied predominantly on recurrent neural networks (RNNs), particularly long short-term memory (LSTM) networks, to model the temporal dynamics of IMU signals. Musci et al.~\cite{musci2019online} proposed an RNN architecture based on LSTM cells deployed on a smartwatch for online fall detection, showing the feasibility of real-time sequential modeling from wrist-mounted accelerometers. Subsequent works adopted hybrid CNN-LSTM~\cite{info12100403} architectures to jointly capture local spatial features and long-range temporal dependencies: the CNN layers extract discriminative motion patterns from fixed-length windows, while the LSTM layers model the temporal evolution of those patterns across time steps. Despite their adoption, recurrent architectures carry several limitations. Their sequential processing of hidden states prevents parallelization during training, resulting in high computational cost and slow convergence. Moreover, LSTM-based models are prone to vanishing gradients over long sequences and tend to accumulate error across time steps when the informative event, i.e., the impact phase of a fall, is confined to a small fraction of the input window. These limitations motivated the community to explore attention-based alternatives.

\subsection{Transformer Attention in Fall Detection}

The introduction of self-attention mechanisms~\cite{vaswani2017attention} offered a compelling solution to the limitations of recurrent networks: by computing pairwise interactions among all positions in a sequence in parallel, transformers eliminate the sequential bottleneck and support direct modeling of long-range dependencies. In the fall detection domain, Yhdego~\cite{Yhdego2021} proposed a transformer encoder with Time2Vec positional encoding for wrist-based fall detection, outperforming LSTM and CNN architectures across multiple datasets. Yasmin et al.~\cite{yasmin2025enhancing} conducted a systematic study applying a transformer encoder on accelerometer and gyroscope data, achieving better performance than the CNN-LSTM architecture. 

Zhang et al.~\cite{dscs2024} proposed the Dual-Stream CNN with Self-Attention (DSCS) model, which applies a self-attention module after a three-layer CNN to dynamically weight accelerometer and gyroscope feature vectors.
More recently, Haque et al.~\cite{haque2024experimental} applied a transformer-based framework for real-time activity recognition and fall detection using wearable IMU data, outperforming CNN-LSTM and temporal convolutional network (TCN) baselines. Pradhan et al.~\cite{bdcc10030090} further incorporated squeeze-and-excitation blocks with temporal attention pooling on top of a transformer backbone for wrist-based fall detection, achieving the state-of-the-art performance. 

Despite these advancements, the structural limitations of self-attention, as discussed in Section~\ref{sect:introduction}, namely $\mathcal{O}(T^2 d_k)$ overhead and softmax-induced dilution of discriminative activations, remain unaddressed in all of the above works, motivating the exploration of gating as a structurally aligned alternative for short IMU windows.

\subsection{Gating Mechanisms Beyond Fall Detection}

Gated linear units (GLU), introduced by Dauphin et al.~\cite{glu2017} for language modeling, emerged as a computationally efficient alternative to attention for selective feature refinement. In the time series domain, Liu et al.~\cite{Liu2021GatedTN} proposed Gated Transformer Networks (GTN) for multivariate time series classification, demonstrating that a gating layer merging channel-wise and step-wise transformer achieves competitive accuracy. For HAR specifically, Bfranc et al.~\cite{Bolatov2024GLULA} introduced GLU-based linear attention (LA), which combined gated convolutional networks with branched convolutions and linear attention. The method achieved state-of-the-art performance across four datasets among contemporary models. However, in GLULA, the gating block receives features already processed by a linear attention module. Because softmax attention distributes non-zero weights across all time steps, including uninformative ones, the features entering the gate carry residual activations from background regions. Consequently, the gate cannot fully suppress those regions, as it operates on a representation that attention has already partially activated, rather than on the raw convolutional outputs where the background activations are near zero.

To address this, we propose a Gated-CNN method, which replaces attention with a dedicated gating module, operating directly on contextually projected features. Our method shows that for short fixed-length IMU windows, sigmoid gating provides sufficient and more precise selective suppression of uninformative activations.

\section{Methodology}\label{sect:methodology}
\begin{figure*}
 \centering
 \includegraphics[width=1.0\textwidth]{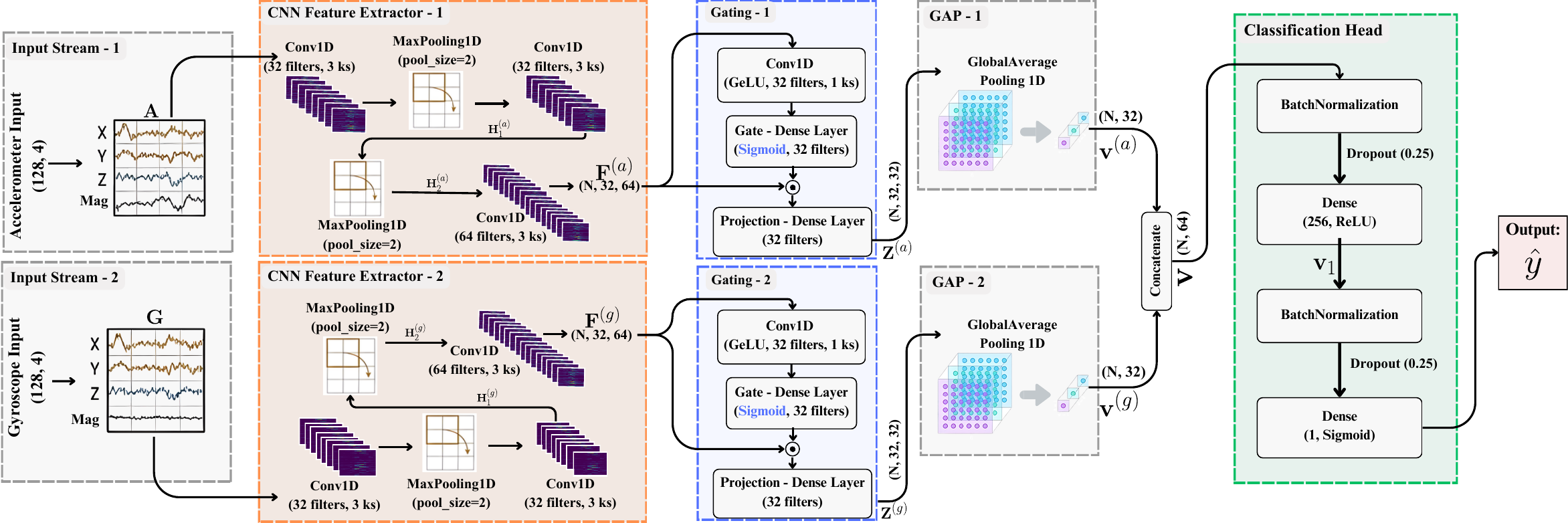}
 \caption{Overview of the proposed dual-stream gated convolutional neural network: Accelerometer $\mathbf{A} \in \mathbb{R}^{128 \times 4}$ and gyroscope $\mathbf{G} \in \mathbb{R}^{128 \times 4}$ inputs are processed through parallel CNN Feature Extractors and Gating modules, followed by Global Average Pooling to yield $\mathbf{v}^{(a)}, \mathbf{v}^{(g)} \in \mathbb{R}^{32}$. The concatenated representation $\mathbf{V} \in \mathbb{R}^{64}$ is passed to the Classification Head to produce the binary output $\hat{y} \in [0, 1]$.}
 \label{fig:cnn_gating_flowchart}
\end{figure*}
This section presents the proposed dual-stream Gated-CNN architecture for sensor-based motion classification. As illustrated in 
Fig.~\ref{fig:cnn_gating_flowchart}, the model processes two parallel IMU input streams, accelerometer and gyroscope, each with four channels ($x$, $y$, $z$, 
and magnitude). These streams are processed through four core components: (1) a CNN Feature Extractor that applies hierarchical 1D convolutions to capture temporal motion patterns, (2) a Gating module that suppresses irrelevant activations via a learned soft mask, (3) a Global Average Pooling (GAP) layer that compresses temporal features into a compact descriptor, and (4) a shared 
Classification Head that fuses both stream descriptors and produces a 
binary prediction. Both streams pass through identical layer configurations, but maintain independent, unshared weights, allowing each modality to learn its own discriminative representations.

\subsection{Input Streams}

The model begins by reading raw tri-axial sensor measurements from each stream. 
For the accelerometer and gyroscope, the signal vectors (SV) at time step $t$ 
are defined as:
\begin{equation}
\begin{aligned}
    \mathbf{SV}^{(a)}_t = [x^{a}_t, y^{a}_t, z^{a}_t] \in \mathbb{R}^{3}, \\
    \mathbf{SV}^{(g)}_t = [x^{g}_t, y^{g}_t, z^{g}_t] \in \mathbb{R}^{3},
\end{aligned}
\end{equation}
where $x^{a}_t$, $y^{a}_t$, $z^{a}_t \in \mathbb{R}$ and $x^{g}_t$, $y^{g}_t$, 
$z^{g}_t \in \mathbb{R}$ denote the measurements ($a$) of the accelerometer and gyroscope ($g$) along the $x$, $y$, and $z$ axes at time step $t$, respectively. To provide a summary of the overall signal intensity, the magnitude feature is computed as the Euclidean norm of each SV:
\begin{equation}
\begin{aligned}
    \text{mag}^{(a)}_t = \sqrt{(x^{a}_t)^2 + (y^{a}_t)^2 + (z^{a}_t)^2} \in \mathbb{R}, 
    \\
    \text{mag}^{(g)}_t = \sqrt{(x^{g}_t)^2 + (y^{g}_t)^2 + (z^{g}_t)^2} \in \mathbb{R},
\end{aligned}
\end{equation}
where $\text{mag}^{(a)}_t$ and $\text{mag}^{(g)}_t \in \mathbb{R}$ are the scalar magnitudes of the accelerometer and gyroscope streams at time step $t$, respectively. Each magnitude is then horizontally concatenated with its corresponding SV to form the Signal Magnitude Vector (SMV)~\cite{10557918}:
\begin{equation}
\begin{aligned}
    \mathbf{SMV}^{(a)}_t = [\mathbf{SV}^{(a)}_t, \text{mag}^{(1)}_t] \in \mathbb{R}^{4}, 
    \\
    \mathbf{SMV}^{(g)}_t = [\mathbf{SV}^{(g)}_t, \text{mag}^{(2)}_t] \in \mathbb{R}^{4},
\end{aligned}
\end{equation}
where $\mathbf{SMV}^{(a)}_t$ and $\mathbf{SMV}^{(g)}_t \in \mathbb{R}^{4}$ are the four-channel input vectors for the accelerometer and gyroscope at time step $t$, capturing both directional motion components and total signal energy. Each stream is then organized into a windowed tensor by stacking $W$ consecutive 
SMV vectors along the temporal axis, where $N$ is the batch size and $W$ is the 
number of time steps in the sliding window, yielding the final input tensors:
\begin{equation}
\begin{aligned}
    \mathbf{A} = \left[\mathbf{SMV}^{(a)}_1, \ldots, \mathbf{SMV}^{(a)}_W
    \right] \in \mathbb{R}^{N \times W \times 4}, \\
    \mathbf{G} = \left[\mathbf{SMV}^{(g)}_1, \ldots, \mathbf{SMV}^{(g)}_W
    \right] \in \mathbb{R}^{N \times W \times 4},
\end{aligned}
\end{equation}
where $\mathbf{A} \in \mathbb{R}^{N \times W \times 4}$ and 
$\mathbf{G} \in \mathbb{R}^{N \times W \times 4}$ denote the accelerometer 
and gyroscope input tensors, respectively, which are subsequently fed into the 
dual-stream network.

\subsection{CNN Feature Extractors}

Each stream passes through its own CNN Feature Extractor to extract contextually projected features. This module is composed of successive 1D convolutional and pooling layers that extract temporal motion features at increasing levels of abstraction. The extractor for Stream 1 (top row from left to right in Fig.~\ref{fig:cnn_gating_flowchart}) takes $\mathbf{A} \in \mathbb{R}^{N \times W \times 4}$ as input, and the extractor for Stream 2 (bottom row from left to right in Fig.~\ref{fig:cnn_gating_flowchart}) takes $\mathbf{G} \in \mathbb{R}^{N \times W \times 4}$ as input. Both extractors 
follow identical layer configurations but maintain independent, unshared weights. 
For notational compactness, the following equations use $\mathbf{I}^{(s)}$ to 
denote the input of stream $s$, where $s \in \{a, g\}$, such that 
$\mathbf{I}^{(a)} = \mathbf{A}$ and $\mathbf{I}^{(g)} = \mathbf{G}$.

In the first stage, a 1-dimensional convolution (Conv1D) layer with $f_1 = 32$ filters, kernel size $ks = 3$, Rectified Linear Unit (ReLU) activation, and same padding (output size equals input size) is applied to $\mathbf{I}^{(s)}$, followed by a MaxPooling1D (MaxPool) layer with pool size $p = 2$, halving the temporal dimension and producing the intermediate feature map $\mathbf{H}^{(s)}_1 \in \mathbb{R}^{N \times \frac{W}{2} \times 32}$:

\begin{equation}
    \mathbf{H}^{(s)}_1 = \text{MaxPool}_{p=2}\!\left( \text{ReLU}\!\left(\text{Conv1D}_{f_1=32,\, ks=3}\!\left(\mathbf{I}^{(s)}\right) \right)\right),  
\end{equation}

In the second stage, another Conv1D layer with $f_2 = 32$ filters, kernel size 
$ks = 3$, and ReLU activation is applied to $\mathbf{H}^{(s)}_1$, followed by a 
second MaxPooling1D layer with pool size $p = 2$, further halving the temporal 
dimension and producing $\mathbf{H}^{(s)}_2 \in \mathbb{R}^{N \times \frac{W}{4} 
\times 32}$:
\begin{equation}
    \mathbf{H}^{(s)}_2 = \text{MaxPool}_{p=2}\!\left( \text{ReLU}\!\left(\text{Conv1D}_{f_2=32,\, ks=3}\!\left(\mathbf{H}^{(s)}_1\right) \right)\right),
\end{equation}
In the third stage, a final Conv1D layer with $f_3 = 64$ filters and kernel size  $ks = 3$ is applied to $\mathbf{H}^{(s)}_2$, expanding the feature depth and producing the final feature map $\mathbf{F}^{(s)}$:
\begin{equation}
    \mathbf{F}^{(s)} = \text{ReLU}\!\left(\text{Conv1D}_{f_3=64,\, k=3}\!\left(
    \mathbf{H}^{(s)}_2\right)\right) \in \mathbb{R}^{N \times \frac{W}{4} \times 64},
\end{equation}
where $\mathbf{F}^{(a)} \in \mathbb{R}^{N \times \frac{W}{4} \times 64}$ and 
$\mathbf{F}^{(g)} \in \mathbb{R}^{N \times \frac{W}{4} \times 64}$ are the 
extracted feature tensors for the accelerometer and gyroscope streams, 
respectively, which are subsequently passed to the Gating module.

\subsection{Gating Module}

The gating module represents the most critical component of the proposed architecture, as it enables each stream to selectively emphasize 
informative temporal features and suppress irrelevant activations before fusion. The feature tensor $\mathbf{F}^{(s)} \in \mathbb{R}^{N \times 
\frac{W}{4} \times 64}$ of each stream is passed through a dedicated Gating module that applies a learned soft mask (i.e., a continuous, learnable filter) via a sigmoid-activated gate, inspired by GLU.

First, a Conv1D layer with $d = 32$ filters, kernel size $k = 1$, and Gaussian Error Linear Unit (GeLU) activation is applied to $\mathbf{F}^{(s)}$, producing the contextual projection $\mathbf{U}^{(s)} \in \mathbb{R}^{N \times \frac{W}{4} \times 32}$:
\begin{equation}
    \mathbf{U}^{(s)} = \text{GeLU}\!\left(\text{Conv1D}_{d=32,\, k=1}\!\left( \mathbf{F}^{(s)}\right)\right),
\end{equation}
where $\Phi(z)$ is the cumulative distribution function of 
the standard normal distribution. In parallel, a Dense layer with $d = 32$ 
units and sigmoid activation is applied to $\mathbf{U}^{(s)}$ to produce the 
gate tensor $\mathbf{\Gamma}^{(s)} \in \mathbb{R}^{N \times \frac{W}{4} \times 
32}$, whose values lie in $(0, 1)$ and control the contribution of each feature:
\begin{equation}
    \mathbf{\Gamma}^{(s)} = \sigma\!\left(\text{Dense}_{32}\!\left(\mathbf{U}^{(s)} \right)\right),
\end{equation}
where $\sigma(z) = \frac{1}{1 + e^{-z}}$ is the sigmoid activation function. 
The gated output $\tilde{\mathbf{F}}^{(s)} \in \mathbb{R}^{N \times \frac{W}{4} 
\times 32}$ is then computed via element-wise multiplication of the projection 
$\mathbf{U}^{(s)}$ and the gate $\mathbf{\Gamma}^{(s)}$:
\begin{equation}
    \tilde{\mathbf{F}}^{(s)} = \mathbf{U}^{(s)} \odot \mathbf{\Gamma}^{(s)},
\end{equation}
where $\odot$ denotes element-wise multiplication. Finally, a linear 
Dense layer with $d = 32$ units projects $\tilde{\mathbf{F}}^{(s)}$ into the 
refined output $\mathbf{Z}^{(s)} \in \mathbb{R}^{N \times \frac{W}{4} \times 
32}$:
\begin{equation}
    \mathbf{Z}^{(s)} = \text{Dense}_{32}\!\left(\tilde{\mathbf{F}}^{(s)}\right),
\end{equation}
where $\mathbf{Z}^{(a)} \in \mathbb{R}^{N \times \frac{W}{4} \times 32}$ and 
$\mathbf{Z}^{(g)} \in \mathbb{R}^{N \times \frac{W}{4} \times 32}$ are the 
refined feature tensors for the accelerometer and gyroscope streams, 
respectively, which are subsequently passed to the individual Global Average Pooling layers.

\subsection{Global Average Pooling}

Following the gating module, a 1D Global Average Pooling (GAP) 
layer collapses the temporal dimension of the refined feature tensor 
$\mathbf{Z}^{(s)} \in \mathbb{R}^{N \times \frac{W}{4} \times 32}$ into a 
compact fixed-length feature vector. For each stream $s \in \{a, g\}$, the GAP 
operation averages activations across all $T = \frac{W}{4}$ temporal positions:
\begin{equation}
    \mathbf{v}^{(s)} = \frac{1}{T} \sum_{t=1}^{T} \mathbf{Z}^{(s)}_t \in  \mathbb{R}^{N \times 32},
\end{equation}
where $\mathbf{Z}^{(s)}_t \in \mathbb{R}^{32}$ is the feature vector at temporal position $t$, $T = \frac{W}{4}$ is the total number of temporal positions in $\mathbf{Z}^{(s)}$, and $\mathbf{v}^{(s)} \in \mathbb{R}^{N \times 32}$ is the resulting vector for stream $s$. This produces two fixed-length descriptors $\mathbf{v}^{(a)} \in \mathbb{R}^{N \times 32}$ and $\mathbf{v}^{(g)} \in \mathbb{R}^{N \times 32}$ for the accelerometer and gyroscope streams, respectively, which are subsequently passed to the Feature Fusion step.

\subsection{Feature Fusion}

The two stream descriptors produced by the GAP layers are concatenated along the feature axis to form a unified multimodal representation:
\begin{equation}
    \mathbf{v} = \left[\mathbf{v}^{(a)} \;\|\; \mathbf{v}^{(g)}\right] \in  \mathbb{R}^{N \times 64},
\end{equation}
where 
$[\cdot \| \cdot]$ denotes concatenation along the feature dimension, and $\mathbf{v} \in \mathbb{R}^{N \times 64}$ is the resulting fused 
representation. This concatenation preserves the distinct contributions of each 
modality, allowing the classification head to jointly reason over accelerometer 
and gyroscope features without discarding sensor-specific structure. 

\subsection{Classification Head}

The fused representation $\mathbf{v} \in \mathbb{R}^{N \times 64}$ is passed 
through a fully connected classification head composed of an interleaved batch 
normalization, dropout, and dense layers. In the first block, batch normalization and dropout are applied to $\mathbf{v}$ before a Dense layer with $d = 256$ units and ReLU activation, producing $\mathbf{v}_1 \in \mathbb{R}^{N \times 256}$:
\begin{equation}
    \mathbf{v}_1 = \text{ReLU}\!\left(\mathbf{\Theta}_1 \cdot  \text{Dropout}_{p=0.25}\!\left(\text{BatchNorm}(\mathbf{v})\right) +  \mathbf{b}_1\right),
\end{equation}
where $\mathbf{\Theta}_1 \in \mathbb{R}^{256 \times 64}$ is the learnable weight matrix, $\mathbf{b}_1 \in \mathbb{R}^{256}$ is the bias vector, 
$\text{BatchNorm}(\cdot)$ normalizes activations to zero mean and unit variance 
across the batch, $\text{Dropout}_{p}(\cdot)$ randomly zeros $p \times 100\%$ 
of activations during training, and $\text{ReLU}(z) = \max(0, z)$. In the second block, the same sequence of batch normalization and dropout is applied to $\mathbf{v}_1$ before the final output Dense layer with a single unit and sigmoid activation, producing the predicted probability $\hat{y}$:
\begin{equation}
    \hat{y} = \sigma\!\left(\mathbf{\Theta}_2 \cdot \text{Dropout}_{p=0.25}\!\left(\text{BatchNorm}(\mathbf{v}_1)\right) + b_2\right) \in \mathbb{R}^{N \times 1},
\end{equation}
where $\mathbf{\Theta}_2 \in \mathbb{R}^{1 \times 256}$ is the output weight 
vector, $b_2 \in \mathbb{R}$ is the scalar bias term, and $\sigma(z) = 
\frac{1}{1+e^{-z}}$ is the sigmoid activation function. The output $\hat{y} \in 
[0, 1]$ represents the predicted probability of the positive class, enabling 
binary classification.

\section{Implementation Details}\label{sect:impl_details}

\subsection{Datasets}\label{subsec:dataset}
We evaluated the proposed method using five datasets that collected triaxial accelerometer and gyroscope data from wrist-mounted sensors: UP-Fall~\cite{upfall}, WEDA-FALL~\cite{wedafall}, UMAFall~\cite{UMAFall2017}, FallAllD~\cite{fallalld}, and SmartFallMM~\cite{smartfallmm2025}.

UP-Fall~\cite{upfall} includes data from 17 participants (aged 18--24), performing 11 activities: six activities of daily living (ADLs) (walking, standing, sitting, picking up an object, jumping, lying) and five fall types (forward using hands, forward using knees, backward, sideward, sitting in an empty chair). Data were captured using MbientLab MetaSensor IMUs at 100~Hz.

WEDA-FALL~\cite{wedafall} includes data from 14 adults (aged 20--46), performing 11 ADLs (e.g., walking, jogging, crouching, clapping) and eight fall types (e.g., forward, lateral, and backward falls while walking or sitting). Data were captured using a consumer-grade Fitbit Sense smartwatch at multiple sampling rates, and for this work, we used the recordings downsampled to 40~Hz.

UMAFall~\cite{UMAFall2017} includes data from 17 participants (aged 18--55), performing eight ADLs (e.g., walking, jogging, sitting down and getting up, climbing stairs) and three fall types (forward, backward, lateral). Data were captured using Texas Instruments SensorTag motes at 20~Hz.

FallAllD~\cite{fallalld} includes data from 15 participants (aged 21--53), performing 44 ADL types and 35 fall types covering all possible directions, causes (slip, trip, syncope, loss of balance). Data were captured using custom RF-TRACK data-loggers at multiple sampling rates, and for this work, we used the recordings downsampled to 40~Hz.

SmartFallMM~\cite{smartfallmm2025} includes data from 30 young adults (aged 18--35), performing 14 activities: nine ADLs (drinking water, picking up objects, putting on a jacket, sweeping, hand washing, waving, walking, sitting, standing) and five fall types (forward, backward, left, right, rotational). The dataset was collected under Institutional Review Board (IRB\#) approvals 7846 and 9461, and data were captured from a commodity Google Pixel Watch 3 at 32~Hz.

\subsection{Data Segmentation}\label{subsect:data_seg}
We employ a Leave-One-Subject-Out Cross-Validation (LOSO-CV) protocol, where in each fold, one subject is designated as the test set, one subject is selected as a validation set for hyperparameter tuning and early stopping, and the remaining subjects constitute the training set. This subject-level partition prevents data leakage and enables robust cross-subject generalization.

We segment the time-series signals into fixed-length overlapping windows of 4 seconds with a stride of 10 samples across all datasets. Given dataset-specific sampling rates, this yields different window sizes per dataset; for instance, for SmartFallMM (32~Hz), this setting produces windows of 128 samples.
We formulate fall detection as a binary classification problem at the window level, where each window containing a fall event is assigned label 1, and windows containing only ADLs are assigned label 0.

The resulting class distribution is inherently imbalanced, with ADL samples constituting approximately 60\% and fall samples approximately 40\% of the total data. To mitigate this effect, we incorporate class weights into the loss function, computed as: $w_c = {N}/{K \cdot N_c}$, where $N$ is the total number of training samples, $K$ is the number of classes, and $N_c$ is the number of samples in class $c$. This assigns a higher penalty to misclassified fall events during training optimization.

\subsection{Training and Evaluation Protocol}\label{subsec:trai_eval_conf}
We train the proposed model end-to-end using the Adam optimizer with binary cross-entropy loss, for up to 250 epochs with a batch size of 32. We apply early stopping with a patience of 10 epochs, monitoring the validation loss, and save the best model checkpoint per fold. We implement all experiments in Python 3.12 with TensorFlow 2.19, using a cluster of eight NVIDIA RTX A5000 GPUs, each with 24~GB of memory.

For results, we report the macro-averaged F1-score as the primary evaluation metric across all experiments. The F1-score is computed as the harmonic mean of precision and recall: $\text{F1-score} = 2 \times (\text{Precision} \times \text{Recall}) / (\text{Precision} + \text{Recall})$. This metric provides a balanced measure of classification performance, as any drop in either precision or recall directly reduces the F1-score.

\section{Results and Discussion}\label{sect:results_discussion}
\subsection{Attention vs Gating Across All Datasets}\label{subsect:attn_vs_gate}
We first compare the proposed Gated-CNN model (see Fig.~\ref{fig:cnn_gating_flowchart}) against a dual-stream Transformer baseline to evaluate whether a gating mechanism offers advantages over standard attention-based temporal modeling. Both models share an identical classification head and the same dual-stream input structure, processing accelerometer and gyroscope streams independently before concatenation. However, they differ in their temporal feature extraction backbone: the Transformer baseline employs two independent multi-head self-attention branches with layer normalization and feed-forward sublayers. The two branches are independent, however identical in architecture, each consisting of 2 Transformer layers with 4 attention heads, an embedding dimension of 64, and a dropout rate of 0.3. Table~\ref{tab:model_comparison} reports the means of mean F1-scores across all five datasets over ten LOSO-CV folds.
\begin{table}[ht]
\centering
\renewcommand{\arraystretch}{1.20}
\caption{Macro-averaged F1-score (\%) across datasets. Results are reported as mean $\pm$ std over ten LOSO-CV folds. $\dagger$ denotes statistical significance over the Transformer baseline ($p < 0.05$, paired t-test).}
\label{tab:model_comparison}
    \begin{tabular}{lcc}
        \toprule
        \multirow{2}{*}{\textbf{Dataset}} & \multicolumn{2}{c}{\textbf{Dual Stream Models}} \\
        \cmidrule(lr){2-3}
         & \textbf{Transformer} & \textbf{Gated-CNN (Ours)} \\
        \midrule
        SmartFallMM & $85 \pm 4.40$ & $\mathbf{93} \pm 3.3^\dagger$ \\ 
        WEDA-Fall   & $88 \pm 6.20$              & $\mathbf{93} \pm 4.53$ \\ 
        FallAllD    & $75 \pm 8.31$              & $\mathbf{90} \pm 5.18$ \\ 
        UMAFall     & $83 \pm 5.93$              & $\mathbf{91} \pm 4.63$ \\ 
        UP-Fall     & $78 \pm 3.06$              & $\mathbf{90} \pm 1.94$ \\
        \bottomrule
    \end{tabular}
\end{table}

The results show that the Gated-CNN model consistently outperforms the Transformer baseline across all datasets, achieving F1-score improvements of \Plus8\%, \Plus5\%, \Plus15\%, \Plus8\%, and \Plus12\% on SmartFallMM, WEDA-Fall, FallAllD, UMAFall, and UP-Fall datasets, respectively. The Gated-CNN also exhibits lower standard deviation (std) across folds, indicating more stable generalization across subjects. The largest improvement is observed on FallAllD ($75 \pm 8.31$ vs. $90 \pm 5.18$), which contains the most diverse fall types, suggesting that the gating mechanism is more effective at capturing complex temporal patterns. To assess statistical significance, we also conducted a paired t-test~\cite{tTest1908} between the two models on the SmartFallMM dataset, obtaining $p = 0.0001$, which confirms that the performance gain of the Gated-CNN over the Transformer is statistically significant ($p < 0.05$). This outcome indicates that the improvement of \Plus8\% by our model is very unlikely to have occurred by chance.

\begin{figure}
 \centering
 \includegraphics[width=0.99\textwidth]{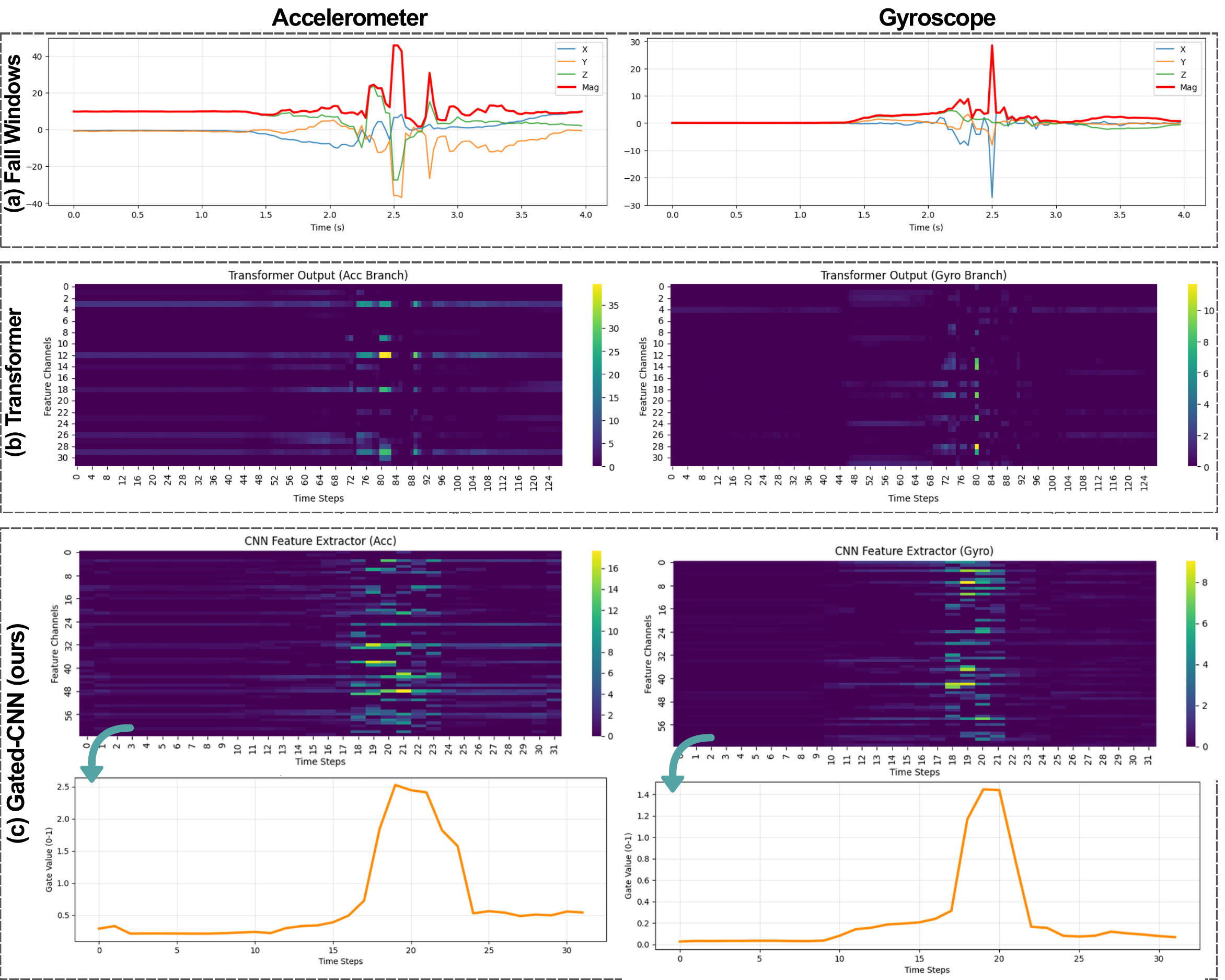}
 \caption{Visualization of feature maps of the Transformer and Gated-CNN on a representative fall window. (a) Raw accelerometer and gyroscope signals with magnitude channel showing a fall impact spike at $\sim$2.5~s. (b) Transformer output feature maps showing sparse activations across both branches. (c) Gated-CNN feature maps with gate activation curves, showing dense activations and a sharp gate response aligned with the impact region.}
 \label{fig:feature_maps}
\end{figure}

To provide qualitative insight into the performance gap observed in Table~\ref{tab:model_comparison}, Fig.~\ref{fig:feature_maps} visualizes the internal representations of both models on a representative fall window from the SmartFallMM dataset. Row~(a) shows the raw accelerometer and gyroscope signals, where a sharp transient spike in the magnitude channel around 2.5~s marks the fall impact. Row~(b) shows the Transformer feature maps, where activations are sparse, scattered, and diluted across both branches. This reflects a fundamental limitation of global self-attention on short windows: by attending uniformly across all time steps, the Transformer fails to preserve the sharp, localized nature of the impact signal. Row~(c) shows the Gated-CNN producing richer and more distributed feature activations across both streams. Critically, the gate activation curves reveal that the model explicitly suppresses background regions, with gate values remaining near-zero before and after the impact and rising sharply to 2.5 and 1.4 in the accelerometer and gyroscope streams, respectively, precisely between time steps 18--25. This selective element-wise temporal control makes the Gated-CNN inherently more suitable for short-window fall detection, where the discriminative signal is brief and easily overwhelmed by pre- and post-impact noise, directly explaining the $+8\%$ F1-score improvements over the Transformer in Table~\ref{tab:model_comparison}.

\subsection{Comparison Across Different Models}\label{subsec:comp_models}
To compare the proposed Gated-CNN against alternative dual-stream architectures, we selected Awatif et al.~\cite{yasmin2025enhancing}, Pradhan et al.~\cite{bdcc10030090}, DSCS~\cite{dscs2024}, and a dual-stream LSTM model~\cite{info12100403}. We trained each model from scratch and evaluated them under the same LOSO-CV protocol and dataset split, using $\mathbf{A} \in \mathbb{R}^{N \times W \times 4}$ and $\mathbf{G} \in \mathbb{R}^{N \times W \times 4}$ as inputs from the SmartFallMM dataset (as described in Section~\ref{sect:methodology}). Table~\ref{tab:arch_comparison} summarizes the mean F1-scores and standard deviations of all evaluated architectures. 

\begin{table}[!htb]
\caption{Performance comparison (mean F1-score \% $\pm$ std) of the proposed Gated-CNN against dual-stream architectures on the SmartFallMM dataset over ten LOSO-CV folds. $\Delta$F1-score denotes the difference ($\text{+}$/$\text{-}$) relative to the proposed model. Bold values indicate the highest performance.}
\label{tab:arch_comparison}
\centering
\renewcommand{\arraystretch}{1.25}
\begin{tabular}{lcc}
    \toprule
    \textbf{Architecture} & \textbf{F1 Score} & \textbf{$\Delta$F1-score} \\
    \midrule
    \makecell[l]{Gated-CNN (Ours)}
        & \textbf{93 $\pm$ 3.3}
        & --- \\
    
    \makecell[l]{Dual-Stream LSTM~\cite{info12100403}}
        & 88 $\pm$ 3.84
        & $\text{-}$4 \\
    
    \makecell[l]{Pradhan et al.~\cite{bdcc10030090}}
        & 85 $\pm$ 4.50
        & $\text{-}$8 \\
    
    \makecell[l]{Awatif et al.~\cite{yasmin2025enhancing}}
        & 78 $\pm$ 4.66
        & $\text{-}$15 \\
    
    \makecell[l]{DSCS~\cite{dscs2024}}
        & 73 $\pm$ 12.53
        & $\text{-}$20 \\
    \bottomrule
\end{tabular}
\end{table}

The Dual-Stream LSTM~\cite{info12100403} achieved 88.84\% ($\pm$3.84), a gap of 4.16\% below the proposed model, with a low standard deviation indicating consistent but lower performance across subjects. Pradhan et al.~\cite{bdcc10030090}, despite incorporating Kalman-based sensor fusion and transformer self-attention with Squeeze-and-Excitation and Temporal Attention Pooling, achieved 85\% ($\pm$4.50), suggesting that the additional preprocessing and architectural complexity did not compensate for the inductive bias provided by the gating mechanism. 
Awatif et al.~\cite{yasmin2025enhancing} achieved 78\% ($\pm$4.66), where the performance gap is attributed to their single-stream transformer architecture that processes concatenated accelerometer and gyroscope inputs, which limits the model's ability to learn modality-specific representations compared to the proposed dual-stream gating design.
The DSCS method~\cite{dscs2024} achieved the lowest mean F1-score of 73\% ($\pm$12.53), with high variance indicating poor generalization under the cross-subject evaluation protocol. Overall, regardless of how attention layers were organized, whether as self-attention, channel attention, or temporal attention pooling, none of the attention-based architectures outperformed the proposed Gated-CNN, suggesting that the sigmoid gating mechanism provides a more effective and consistent form of feature selection for wrist-worn IMU-based fall detection under subject-independent evaluation.

\subsection{Feature Importance Analysis}\label{subsec:feature_importance}
\begin{figure}
 \centering
 \includegraphics[width=0.8\textwidth]{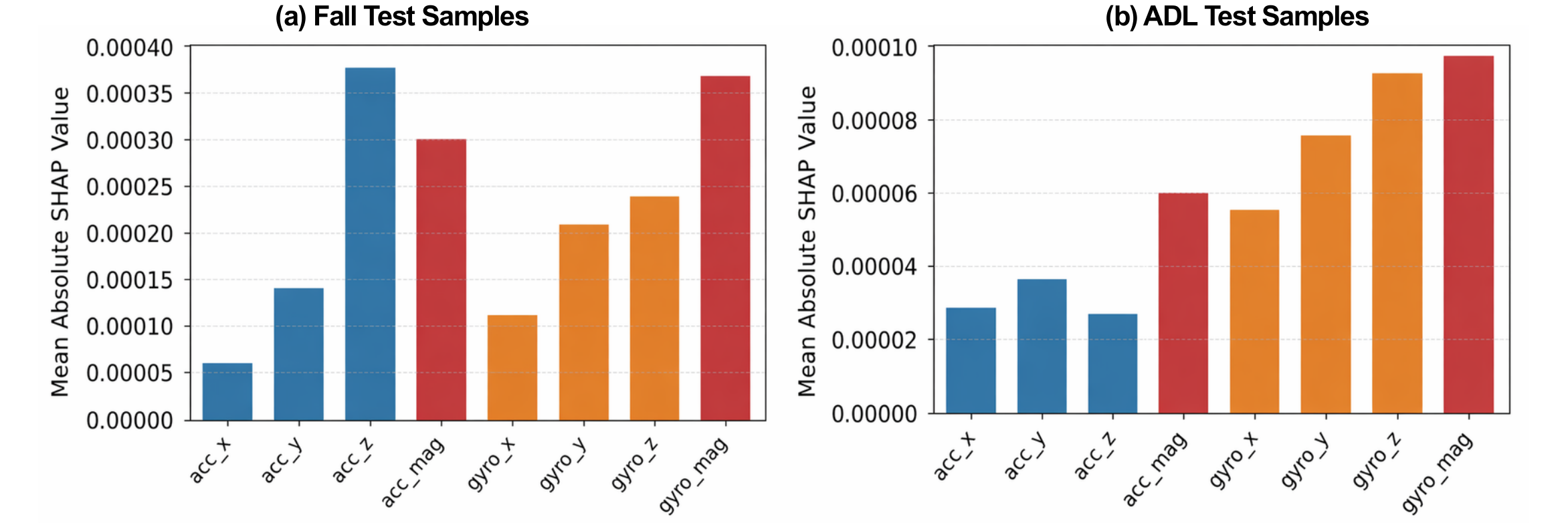}
 \caption{SHAP-based feature importance for the Gated-CNN model on the test set of SmartFallMM dataset~\cite{smartfallmm2025}, reported separately for (a) fall samples and (b) ADL samples. Blue, orange, and red bars show accelerometer, gyroscope, and magnitudes of both channels, respectively.}
 \label{fig:xai_feats_impt}
\end{figure}

To interpret the learned feature importance of the Gated-CNN, we apply SHapley Additive exPlanations (SHAP)~\cite{shap2017}, a post-hoc explainability method grounded in cooperative game theory. SHAP assigns each input feature a contribution value that reflects its marginal impact on the model's output, computed by averaging over all possible feature subsets. We select SHAP specifically because it is model-agnostic and can naturally handle the dual-stream input structure of our model, attributing importance independently to each accelerometer and gyroscope channel. Figure~\ref{fig:xai_feats_impt} reports the mean absolute SHAP values for each input channel, computed separately for fall and ADL test samples.

For fall samples (Figure~\ref{fig:xai_feats_impt}a), the vertical accelerometer axis (\texttt{acc\_z}) and the accelerometer magnitude (\texttt{acc\_mag}) are the most discriminative features, with mean absolute SHAP values of $3.8 \times 10^{-4}$ and $3.0 \times 10^{-4}$, respectively. The gyroscope magnitude (\texttt{gyro\_mag}) also contributes strongly ($3.7 \times 10^{-4}$), indicating that the rotational dynamics during a fall are highly informative. For ADL samples (Figure~\ref{fig:xai_feats_impt}b), the gyroscope channels dominate, particularly \texttt{gyro\_mag} ($9.8 \times 10^{-5}$) and \texttt{gyro\_z} ($9.3 \times 10^{-5}$), while accelerometer channels contribute relatively less. This divergence in feature importance between fall and ADL samples suggests that the model learns distinct discriminative strategies for each class: relying primarily on vertical acceleration and magnitude for fall detection, and on rotational signals for ADL recognition. Notably, the appended magnitude channels (\texttt{acc\_mag} and \texttt{gyro\_mag}) consistently rank among the top contributors in both classes, validating our design choice of augmenting the raw triaxial signals with a magnitude feature.

\subsection{Computational Cost}\label{subsec:compute_cost}
In this section, we analyze the computational cost of all evaluated architectures in terms of parameter counts in thousands (K), floating point operations (FLOPs) in millions (M), and inference time per batch on a cluster of eight NVIDIA RTX A5000 GPUs in milliseconds (ms). Table~\ref{tab:compute_breakdown} summarizes the results obtained using the SmartFallMM dataset.
\begin{table}[!htb]
\caption{Comparison of computational cost in terms of parameter counts, FLOPs, and inference times measured with batch size 32 on NVIDIA RTX A5000 GPUs. Preprocessing times are measured per 128-sample window on the CPU. Bold values show the lowest computational cost.}
\label{tab:compute_breakdown}
\centering
\renewcommand{\arraystretch}{1.25}
\begin{tabular}{lccc}
    \toprule
    \textbf{Method}
    & \textbf{\makecell[c]{Params\\(K)}}
    & \textbf{\makecell[c]{FLOPs\\(M)}}
    & \textbf{\makecell[c]{Inference\\(ms/batch)}} \\
    \midrule
     Gated-CNN (Ours) & \textbf{31} & 3.8 & \textbf{2.8} \\ 
     Pradhan et al.~\cite{bdcc10030090} & 42 & 2.8 & 3.0  \\ 
     Awatif et al.~\cite{yasmin2025enhancing} & 32 & \textbf{2.7} & 3.7  \\ 
     LSTM~\cite{info12100403} & 202 & 66.5 & 6.4  \\
     DSCS~\cite{dscs2024} & 68 & 64.50 & 3.8  \\
    \bottomrule
\end{tabular}
\end{table}

Results show that our method is the most lightweight model, with only 31K parameters, and achieves the fastest inference time of 2.8~ms per batch. 
The Dual-Stream LSTM~\cite{info12100403} is the most computationally expensive baseline, with 202K parameters and 66.5M FLOPs, over $6\times$ the parameters and $17\times$ the FLOPs of the proposed model, while also being the slowest at 6.4~ms per batch, making it impractical for real-time inference on wearable hardware. 
Pradhan et al.~\cite{bdcc10030090}, and Awatif et al.~\cite{yasmin2025enhancing} are more comparable in parameter count at 42K and 32K, respectively, and both exhibit higher inference times than the proposed model. The DSCS method~\cite{dscs2024} achieves the second largest FLOPs at 64.50M, with 68K parameters and 3.8~ms inference time compared to the proposed model. Overall, the Gated-CNN achieves the best trade-off between predictive performance and computational efficiency.

\subsection{Ablation Study}\label{subsec:ablation_study}

To assess the contribution of each architectural component in the proposed Gated-CNN model, we conducted four ablation experiments (T1--T4), systematically removing or replacing individual modules while keeping the remaining pipeline fixed (see Fig.~\ref{fig:cnn_gating_flowchart}). The configuration of each variant is detailed in Table~\ref{tab:ablation_config}, and the resulting F1-score distributions obtained over ten folds of LOSO-CV are summarized in Fig.~\ref{fig:failure_example}.
\begin{table}[ht]
\centering
\renewcommand{\arraystretch}{1.25}
\caption{Configurations of ablation tests, where \ding{51} indicates the component is present, \ding{55} indicates it was removed, and a text entry indicates it was replaced or modified.}
    \begin{tabular}{lccccc}
        \toprule
        \textbf{Model Variant} & 
        \textbf{CNN-Block} & 
        \textbf{Gating} & 
        \textbf{GAP} & 
        \textbf{Concat} & 
        \textbf{Head} \\
        \midrule
        Gated-CNN (Ours)   & \ding{51} & \ding{51} & \ding{51} & \ding{51} & \ding{51}  \\ 
        T1: CNN & \ding{51} & \ding{55} & \ding{51} & \ding{51} & \ding{51}  \\ 
        T2: Gating& \ding{55} & \ding{51} & \ding{51} & \ding{51} & \ding{51}  \\ 
        T3: Linear Proj. & Linear Proj. & \ding{51} & \ding{51} & \ding{51} & \ding{51}  \\ 
        T4: SHAP & \ding{51} & \ding{51} & \ding{51} & \ding{51} & \ding{51} \\ 
        T5: Acc/Gyro & \ding{51} & \ding{51} & \ding{51} & \ding{55} & \ding{51} \\ \bottomrule

    \end{tabular}%
\label{tab:ablation_config}
\end{table}
\begin{figure}
 \centering
 \includegraphics[width=0.76\textwidth]{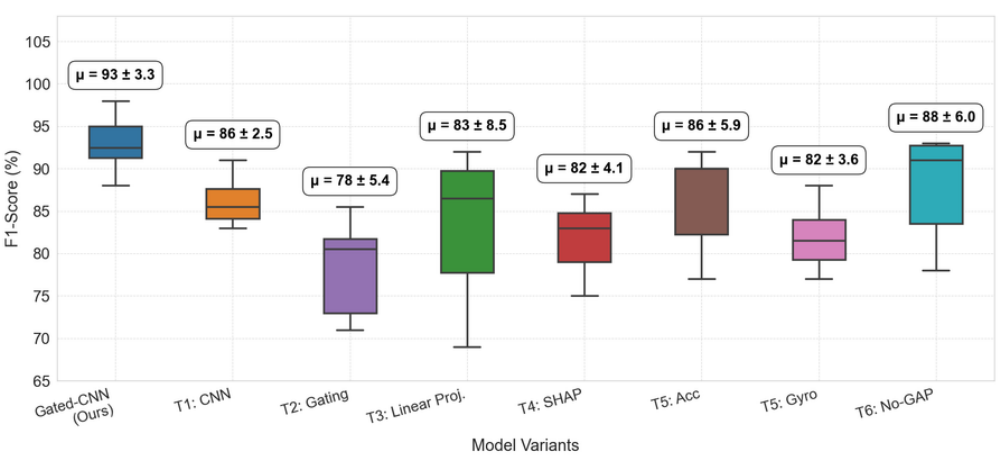}
 \caption{Comparison of F1-score distributions across ablation variants (T1--T4) and the proposed Gated-CNN over LOSO-CV folds. Each box summarizes performance over ten cross-validation folds. The annotated $\mu \pm \sigma$ values denote the mean and standard deviation of F1-scores per variant.}
 \label{fig:ablations}
\end{figure}

\subsubsection{T1: Removing the Gating Module}

In the T1 test, we removed the gating module and connected the CNN feature extractor directly to Global Average Pooling (GAP), bypassing any temporal feature selection. This modification reduced the mean F1-score to 86\% ($\pm$2.54), a 7\% drop relative to the proposed model's 93\% ($\pm$3.17). The consistent decline across all ten folds confirmed that the gating block played a critical role in selectively weighting informative temporal features before pooling. Notably, T1 exhibited the lowest standard deviation among all ablated variants, indicating that the CNN block alone extracted stable and reproducible features across subjects. However, without suppressing non-discriminative activations, the model consistently underperformed, suggesting the critical role of the gating module.

\subsubsection{T2: Removing the CNN Block}

In T2, we removed the CNN feature extractor entirely and fed raw input sequences directly into the gating module. This yielded the most severe degradation among all variants, reducing the mean F1-score to 78\% ($\pm$5.42). The elevated variance further indicated that the gating mechanism alone, without structured convolutional representations, failed to reliably distinguish fall from non-fall events. This outcome highlights that without CNN-derived features as input, the gating mechanism loses its contextual reference for identifying task-relevant patterns, rendering its selective amplification ineffective.

\subsubsection{T3: Replacing CNN Block with Linear Projection}

In T3, we replaced the CNN block with a single layer of CNN for linear projection, while retaining the gating module and the dual-stream structure. The model achieved a mean F1-score of 83.4\% ($\pm$8.53), with the high standard deviation reflecting substantial cross-fold instability. This result indicated that a linear projection could not replicate the hierarchical feature extraction performed by the CNN block, which progressively mapped raw signals through three convolutional layers with $32$, $32$, and $64$ filters, interleaved with max pooling operations, to produce structured temporal representations.

\subsubsection{T4: SHAP-Selected Channel Reduction}

In T4, we retained the full architecture; however, we restricted the input to the two channels most salient according to SHAP analysis, namely \texttt{\_z} and \texttt{\_mag}, for each sensor stream (see Fig.~\ref{subsec:feature_importance}). The model achieved a mean F1-score of 82.1\% ($\pm$4.15), a moderate but consistent decline relative to the proposed model. This result suggested that while \texttt{\_z} and \texttt{\_mag} captured the dominant motion dynamics, the remaining axes encoded complementary information that the model exploited for robust fall detection.

\subsubsection{T5: Single-Stream Model}
In T5, we isolated the contribution of each sensor modality by training two independent single-stream models: one on accelerometer data exclusively (T5-Acc) and one on gyroscope data exclusively (T5-Gyro), with the inter-stream concatenation removed and all remaining components identical to the proposed model. 
T5-Acc achieved a mean F1-score of 86.4\% ($\pm$5.87), and T5-Gyro yielded 81.9\% ($\pm$3.63), both falling notably short of the proposed model's 93\% ($\pm$3.17). The consistent performance gap across both configurations confirms that the two modalities encode complementary information, and that their fusion is essential for robust fall 
detection.

\subsubsection{T6: Removing the GAP Layer}
In T6, we removed the Global Average Pooling (GAP) layer and instead concatenated the two outputs of the gating module directly along the feature axis. This setting produced a flattened $2048$-dimensional vector passed to the classification head, thereby preserving full temporal resolution rather than aggregating it into a global representation. 
T6 achieved a mean F1-score of 88.0\% ($\pm$6.04), a 5\% decline relative to the proposed model. The elevated variance suggests that the classifier struggled to generalise across subjects when exposed to the full uncompressed sequence, whereas GAP's temporal summarisation provided more stable and discriminative representations.

\subsubsection{Overall Comparison}

Across all ablations, every modification to the proposed architecture produced a measurable decline in F1-score, as shown in Fig.~\ref{fig:ablations}. The results collectively reveal that the model's performance stems from the synergistic interaction of its components: structured convolutional representations, selective temporal gating, complementary dual-stream fusion, and global temporal summarisation via GAP. Each module addresses a distinct aspect of the fall detection problem, and their removal or simplification consistently compromised both accuracy and generalisation stability across subjects.

\subsection{Real-time Testing} \label{subsec:realtime_testing}
For real-time evaluation, we deployed the highest-performing model among the ten LOSO-CV folds (F1-score: 98\%) on the watch-based version of the SmartFall application~\cite{SmartFallProject} running on Google Pixel Watch 3 (see~\cite{haque2024experimental} for implementation details). 
Since the model is trained using TensorFlow and saved in the \texttt{.keras} format, we first convert it to TensorFlow Lite (TFLite) for on-device inference, as TFLite is the only inference runtime supported on Wear OS-based smartwatches. The conversion reduces the model size and enables low-latency inference directly on the watch. In this setup, the smartwatch worn on the left wrist continuously collects accelerometer and gyroscope data, which are fed to the on-device TFLite model at 32~Hz for real-time fall prediction. For more details on the Application, please refer to our previous work 

Following the same testing protocol as the SmartFallMM dataset~\cite{smartfallmm2025}, we recruited 12 participants (aged 20--38 years) from the Texas State University community under IRB\# 9461. Each participant performed all 14 activities (9 ADLs and 5 fall types), each repeated five times, for a total of 70 trials per participant. 
For a detailed performance evaluation, Table~\ref{tab:realtime_metrics} reports precision (ratio of true falls among predicted falls), recall (ratio of detected falls among all actual falls), accuracy (overall correct classification rate), and F1-score (harmonic mean of precision and recall) for each participant.
\begin{table}[!htb]
\caption{Real-time per-participant performance (\%) using the SmartFall watch-based application.}
\label{tab:realtime_metrics}
\centering
\renewcommand{\arraystretch}{1.09}
    \begin{tabular}{lcccc}
        \toprule
        \textbf{Participant} & \textbf{Precision} & \textbf{Recall} & \textbf{F1-Score} & \textbf{Accuracy} \\ \midrule
        User A & 89 & 96 & 92 & 94 \\ 
        User B & 96 & 100 & 98 & 99 \\ 
        User C & 83 & 100 & 90 & 95 \\ 
        User D & 100 & 100 & 100 & 100 \\ 
        User E & 100 & 100 & 100 & 100 \\ 
        User F & 100 & 84 & 91 & 93 \\ 
        User G & 100 & 100 & 100 & 100 \\ 
        User H & 100 & 88 & 94 & 96 \\ 
        User I & 100 & 100 & 100 & 100 \\ 
        User J & 100 & 100 & 100 & 100 \\ 
        User K & 100 & 100 & 100 & 100 \\ 
        User L & 100 & 100 & 100 & 100 \\ \hline
        \textbf{Average} & \textbf{97} & \textbf{97} & \textbf{97} & \textbf{98} \\
        \bottomrule
    \end{tabular}
\end{table}

\begin{figure}
 \centering
 \includegraphics[width=0.8\textwidth]{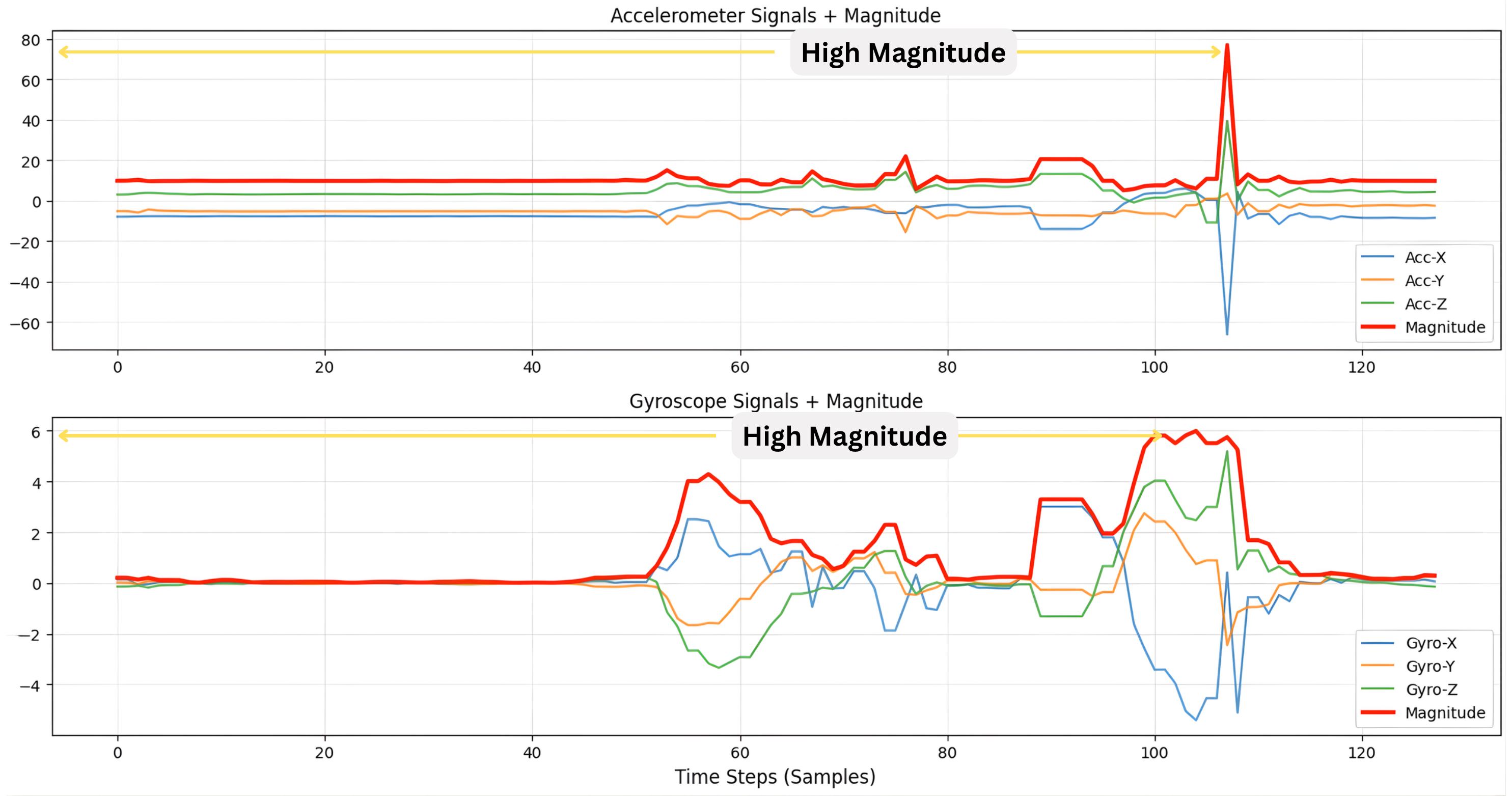}
 \caption{An example of a false positive window from real-time testing: The accelerometer magnitude (top) exhibits sharp peaks exceeding $80$, while the gyroscope magnitude (bottom) sustains elevated angular velocity throughout the window, mimicking the transient impact and rotational dynamics of a fall event.}
 \label{fig:failure_example}
\end{figure}

The model achieves an average F1-score of 97\% and an average accuracy of 98\% across all 12 participants. This result outperforms our prior works~\cite{yasmin2025enhancing, bdcc10030090}, which reported 80--83\% F1-score, representing gains of up to 17\%, representing strong real-time generalization. 
Eight out of twelve participants achieved perfect scores across all metrics, and no false negatives were reported across any participant, confirming that the Gated-CNN reliably detects all fall events in real-time deployment. The observed failures were exclusively false positives, grouped into two recurring activity patterns: (1) \textit{putting on and taking off a jacket}, and (2) \textit{sweeping}. As illustrated in Fig.~\ref{fig:failure_example}, these activities share a common characteristic: they involve sudden or vigorous wrist movements that produce acceleration magnitudes exceeding those present in the SmartFallMM training set (e.g., \texttt{acc\_mag} peaks exceeding $80$), causing the model to misclassify them as falls. This suggests that expanding the training set with high-magnitude ADL samples would further reduce false positives and improve real-world precision.

\section{Conclusion and Future Work}\label{sect:conclusion}
In this paper, we have proposed Gated-CNN, a lightweight dual-stream architecture for smartwatch-based fall detection. The model processes accelerometer and gyroscope streams through (i) independent 1D-CNN feature extractors, followed by (ii) a sigmoid gating module that selectively suppresses uninformative background activations while amplifying fall-discriminative features, (iii) a global average pooling layer that compresses each stream into a compact fixed-length descriptor, and (iv) a shared classification head that fuses both descriptors for binary fall prediction. Rather than relying on self-attention, Gated-CNN employs computationally efficient gating that operates directly on convolutional feature representations.

Offline evaluation across five wrist-mounted IMU datasets under a Leave-One-Subject-Out Cross-Validation protocol showed that the model consistently outperforms Transformer baselines, achieving mean F1-scores between 90\% and 93\% with only 31K parameters and 2.8 ms inference per batch. 
Ablation studies validated the necessity of each architectural component, as removing or simplifying the CNN block, gating module, dual-stream fusion, or GAP consistently degraded both mean F1-score and cross-fold stability. 
Real-time deployment on a Google Pixel Watch 3 and testing across 12 participants further confirmed the practical viability of the approach, yielding an average F1-score of 97\% and an accuracy of 98\% with zero missed falls. 

The observed failures were exclusively false positives caused by high-magnitude daily activities such as putting on a jacket and sweeping, whose peak accelerometer magnitudes exceeded those present in the training data, thereby mimicking the transient impact dynamics of a fall. These failure cases motivate two directions for future work: first, augmenting the training set with high-magnitude daily activity samples to reduce false positives, and second, fine-tuning the model continuously with user-specific feedback data to improve personalized discrimination between vigorous daily activities and genuine fall events.

\section{Resources}
To facilitate reproducibility and further research, we make our trained model and source code publicly available at the following link: \url{https://github.com/txst-cs-smartfall/Gated-CNN-for-Watch-based-Fall-Detection}.

\section*{Acknowledgment}
This research was funded by the National Science Foundation (NSF) under the Smart and Connected Health (SCH) Program grant number 21223749.

\bibliographystyle{unsrtnat}
\bibliography{references}  

\end{document}